\documentclass{article}

\usepackage[final]{neurips_2024}

\usepackage[utf8]{inputenc} % allow utf-8 input
\usepackage[T1]{fontenc}    % use 8-bit T1 fonts
\usepackage{hyperref}       % hyperlinks
\usepackage{url}            % simple URL typesetting
\usepackage{booktabs}       % professional-quality tables
\usepackage{amsfonts}       % blackboard math symbols
\usepackage{nicefrac}       % compact symbols for 1/2, etc.
\usepackage{microtype}      % microtypography
\usepackage{xcolor}         % colors
\usepackage{float}

\title{From Scarcity to Efficiency: Investigating the Effects of Data Augmentation on African Machine Translation}

\author{\normalfont
  Mardiyyah Oduwole, Oluwatosin Olajide, Jamiu Suleiman, Faith Hunja, Busayo Awobade,\\
  Fatimo Adebanjo, Comfort Akanni, Chinonyelum Igwe, Peace Ododo, Promise Omoigui,\\
  Abraham Owodunni, Steven Kolawole \\[1em]
  ML Collective \\
  \texttt{mardiyyah.oduwole@mlcollective.org}
}

\begin{document}
\maketitle

\maketitle

\begin{abstract}
The linguistic diversity across the African continent presents different challenges and opportunities for machine translation. This study explores the effects of data augmentation techniques in improving translation systems in low-resource African languages. We focus on two data augmentation techniques: sentence concatenation with back translation and switch-out, applying them across six African languages. Our experiments show significant improvements in machine translation performance, with a minimum increase of 25\% in BLEU score across all six languages. We provide a comprehensive analysis and highlight the potential of these techniques to improve machine translation systems for low-resource languages, contributing to the development of more robust translation systems for under-resourced languages. 
\end{abstract}

% \begin{keywords}
% Data augmentation for African Languages, switch-out data augmentation for African languages, and sentence concatenation with back translation for African Languages.
% \end{keywords}

\section{Introduction}
\label{sec:intro}

Despite rapid advances in model architectures and the recent surge in large language models, the success of machine translation (MT) systems still fundamentally relies on the availability of extensive parallel corpora. In high-resource languages like English, French, or German, millions of aligned sentence pairs enable models to learn rich linguistic representations and produce high-quality translations \cite{koehn2017six, brants2007large}. \citep{koehn2017six, brants2007large}. However, for many low-resource languages, including several African languages, the acute scarcity of such data creates a significant bottleneck. The lack of sufficient parallel corpora not only hampers the overall performance of MT systems but also limits their ability to capture complex linguistic phenomena, resulting in poorer translation quality compared to high-resource language pairs \citep{sennrich2016edinburgh, haddow2022survey}. Although large language models have been increasingly adopted in various real-world applications \citep{minaee2024large, naveed2023comprehensive, le2023bloom, costa2022no, zhang2023machine}, significant challenges in MT remain. In particular, these models struggle with idiomatic expressions, domain-specific terminology, and the unique linguistic characteristics of low-resource languages.

Data augmentation has emerged as a promising strategy to alleviate the data scarcity problem by generating synthetic training data. Techniques such as switchout and back-translation have improved model robustness by enriching training datasets with controlled variations and additional context. Switchout, for instance, injects controlled noise by randomly replacing words in both source and target sentences during training, helping models generalise better and handle rare or out-of-vocabulary words \citep{wang2018switchout}. Similarly, adding sentence concatenation to the back-translation process has been shown to make synthetic training data more coherent and diverse \citep{kondo2021sentence}. Although these methods were initially proven effective in high-resource scenarios, they hold considerable promise for mitigating the limitations imposed by the scarcity of data in low-resource language pairs.

In this work, we explore how data augmentation techniques can make neural machine translation (NMT) models more robust, context-aware, and capable of handling the subtleties of African languages. Specifically, we investigate non-generative augmentation methods, sentence concatenation with back-translation and switchout using the MaFAND dataset \citep{adelani2022few}. Our study focuses on translating English into four low-resource languages (Swahili, Yoruba, Hausa, and Setswana) and French into two low-resource languages (Wolof and Fon). Our preliminary results suggest that leveraging data augmentation to enrich training data can significantly boost translation quality and coverage for languages that have historically been under-represented in the MT space.

% \textsf{epsfig}.\footnote{See
% \url{http://www.ctan.org/pkg/l2tabu}}
\section{Related work}
\label{sec:lit-review}
\subsection{Data Scarcity challenge in  Low-Resource Languages}
Current machine translation (MT) systems have reached a point where researchers debate whether they can rival human translators in performance \citep{hassan2018achieving, laubli2018has,toral2018attaining, popel2020transforming}. However, these state-of-the-art systems are typically trained on datasets containing tens or even hundreds of millions of parallel sentences. NMT systems, in particular, rely heavily on vast amounts of parallel data for effective training. For example, \citep{koehn2017six} demonstrated in a case study on English-to-Spanish translation that NMT significantly underperforms statistical machine translation (SMT) when trained on fewer than 100 million words. This finding underscores the direct impact of data volume on translation quality, highlighting the challenges of applying NMT to scenarios where such large-scale data is unavailable. Datasets of this magnitude are available only for a limited number of highly resourced language pairs. High-resource pairs such as English and French benefit from decades of curated parallel corpora, ensuring robust training material for MT systems. In stark contrast, the vast majority of global language pairs suffer from extreme data scarcity or, in some cases, a complete absence of parallel data, which poses a major challenge for low-resource languages \citep{koehn2017six, popel2020transforming}. The challenge of acquiring quality parallel data is compounded by the difficulties inherent in large-scale web crawling for low-resource languages. Typical crawling pipelines depend on multiple processing stages and resources such as text preprocessors, bilingual dictionaries, sentence-embedding tools, and preliminary translation systems which may be either unavailable or of substandard quality for low-resource pairs. These challenges highlight the need for alternative strategies to bridge the resource gap for low-resource languages.

\subsection{Data Augmentation Approaches for Neural Machine Translation}
Data augmentation (DA) is a training paradigm that has proven effective across various modalities including computer vision, speech recognition, and natural language processing \citep{park2019specaugment, wang2017effectiveness}. In the context of NMT, DA has emerged as a frontier of research because it offers promising solutions for mitigating the data scarcity challenges inherent in low-resource settings \citep{sennrich2016edinburgh, norouzi2016reward, zhang2016exploiting, fadaee2017data, wang2018switchout, edunov2018understanding, zhang2019regularizing, fadaee2018back}

\subsubsection{Back-Translation}
One of the most well-known data augmentation approaches in NMT is back-translation (BT). In this method, target-language sentences are translated into the source language to generate synthetic parallel data, which is then mixed with the original parallel corpus to retrain the model \citep{sennrich2016edinburgh}. Research has shown that translating target sentences into the source language generally yields better performance compared to the reverse direction. This approach is particularly effective in low-resource NMT, as demonstrated by \citet{chaudhary2019low}, who reported significant quality improvements. Moreover, \citet{edunov2018understanding} provided evidence that back-translation enhances translation performance even on a very large scale while offering benefits in simulated low-resource conditions. Additionally, \citet{kondo2021sentence} have proposed augmenting parallel data by combining back-translation with sentence concatenation, further enriching the training corpus with varied linguistic contexts.

\subsubsection{Advancements in switchout}
Switchout has emerged as a prominent data augmentation technique in NMT due to its simplicity and effectiveness. Unlike more complex approaches that rely on extensive external resources, switchOut operates by randomly replacing words in both the source and target sentences with tokens uniformly sampled from their respective vocabularies. This process injects controlled noise, defined as small, deliberate modifications into training examples while largely preserving their contextual structure. As a result, model robustness is enhanced and overfitting is mitigated \citep{wang2018switchout, sennrich2016edinburgh}. Its computational efficiency makes it particularly attractive in low-resource scenarios where obtaining large parallel corpora is challenging.
Building on the foundational work, subsequent studies have explored refinements and hybrid strategies to further improve the efficacy of switchout. For example, researchers have investigated variants that incorporate semantic or frequency-based constraints into the word replacement process. Furthermore, recent work has demonstrated that combining switchOut with other data augmentation techniques such as back-translation or targeted word substitution can produce synergistic effects, leading to notable improvements in translation quality for low-resource language pairs \citep{fadaee2017data, edunov2018understanding}. These hybrid approaches underscore the potential of switchout not only as a standalone augmentation method but also as a complementary tool within a broader data augmentation framework for NMT.

\subsection{Implications for African Machine Translation}
While previous research has demonstrated the effectiveness of data augmentation techniques such as back-translation, sentence concatenation, and switchout in improving translation performance, these studies have predominantly focused on high-resource language pairs or well-studied low-resource languages. To the best of our knowledge, no prior work has systematically explored the impact of data augmentation on machine translation for the six African language pairs examined in this study. This gap is particularly significant given that African languages often suffer from severe data scarcity and limited representation in existing MT corpora, which impedes the development of robust translation systems.\\
In response to this gap, our work seeks to bridge the divide by evaluating and comparing the performance of back-translation, switchout, and hybrid data augmentation approaches in the context of African machine translation. We hypothesize that these DA techniques, when carefully tailored to the linguistic characteristics and data constraints of African languages, can substantially improve translation quality. By providing empirical evidence on the efficacy of these methods for the targeted language pairs, our study aims to advance the current body of research in African MT and inspire further research into data augmentation strategies for other under-represented languages.

\section{Methodology}
\label{sec:method}
\subsection{Sentence concatenation with back-translation}

Our methodology for refining machine translation models for African languages combines sentence concatenation with back translation (BT) using the mBART model. We begin with back translation—sentences from the original dataset are translated to a target language and then retranslated back to the source language, generating semantically equivalent yet structurally varied text. We integrate this with sentence concatenation, where sentences from the original and back-translated datasets are paired and concatenated. This method was systematically tested at varying degrees of data augmentation, executing experiments with 10\%, 20\%, 30\%, and up to 100\% concatenation rates, where higher rates indicate more extensive dataset modifications. This multifaceted approach not only introduces complex sentence structures but also broadens the training data’s contextual spectrum, vital for teaching the model advanced language patterns crucial for nuanced translation tasks. This series of experiments was meticulously documented to determine the optimal balance for data augmentation while preserving the linguistic integrity essential for accurate machine translation.

\subsection{Switch out}

SwitchOut works by replacing tokens in text data, making small, controlled changes that increase lexical diversity and help the model generalize better during training.
In contrast to generative approaches, switchout replaces tokens with alternatives from either the same language vocabulary (in-lang) or a different language vocabulary (out-lang). This adds to the dataset while keeping its linguistic properties. We tokenized the text data using mBart's tokenizer and applied switchout in 2 operations:
\begin{itemize}
    \itemsep -3pt {} 
    \item \textbf{In-lang switchout:} Here, we replaced tokens with tokens within the same language vocabulary. This operation introduces variations within the linguistic context of the source language while maintaining coherence.
    \item \textbf{Out-lang switchout:} Here, we replaced tokens with tokens sampled from a different language vocabulary. This technique enriches the dataset by introducing cross-linguistic variations, potentially enhancing the model's ability to handle language interactions.
 \end{itemize}
We applied switchout at varying degrees ranging from minimal perturbations to extensive modifications of the randomly shuffled training data. We experimented with switchout rates of 10\%, 20\%, 30\%, 50\%, and 100\%, with higher percentages indicating a greater proportion of tokens subject to replacement. The switchout operations were conducted across multiple the six language pairs.

\section{Experiments} %add citation before MaFAND and mbart model
\label{sec:experiments}
In this section, we provide a comprehensive overview of the experiments conducted to enhance machine translation models for African languages using the MaFAND dataset. The experiments focus on various augmentation techniques aimed at improving translation quality and coverage for low-resource languages. Each technique was evaluated using the mBart model \citet{liu2020multilingual} across six African languages: Swahili, Yoruba, Hausa, Fon, Wolof, and Setswana. The experiments aim to enrich the training data and enhance translation quality.  

\subsection{Setup}   
For our experiments using the MaFAND dataset, we conducted experiments with 2 augmentation techniques: switchout and sentence concatenation with back-translation in 6 African languages paired individually with English and French, as shown in Table\ref{tab:sentenceconcat_results} and Table\ref{tab:switchout_results}.  
Parallel sentences in the dataset for the selected languages typically include the source and target languages, each containing 2100–30782 parallel sentences for modelling. We exclusively tested with the mBART \cite{liu2020multilingual} for our preliminary results. To ensure that our training runs are consistent, we repeated each experiment using three seeds, and the results were averaged. The metrics reported to measure performance are loss and BLEU, as common with NMT systems \cite{zhang2023machine, oh2023data, zhang2024binarized}.

\subsection{Results}
% \label{sec:results}

Table\ref{tab:data augmentation_results} presents the results of our experiments at the best-performing augmentation percentage results for six language pairs, comparing the baseline model with two augmentation techniques: switchout (in-language and out-language) and sentence concatenation with BT—using varying percentages and types of augmentation. The best result for each language pair is highlighted. Our findings indicate that the performance of augmentation techniques varies significantly with the language, the degree of modification, and the type of augmentation applied. For the en-hau pair, in-language switchout at a 50\% augmentation level outperformed both the baseline and the sentence concatenation with backtranslation method. Similarly, for the en-yor pair, out-language switchout at a 30\% augmentation rate yielded the besperformance,  compared to the baseline and sentence concatenation with BT. In contrast, the en-swa pair did not benefit from data augmentation; the baseline model maintained the highest BLEU score with the lowest perplexity. We attribute this to the fact that en-swa had the most parallel data in the MaFAND dataset, which may reduce the marginal improvements offered by augmentation techniques. For en-tsn, while switchout (at 100\% augmentation) improved performance over the baseline, sentence concatenation with backtranslation at a 20\% augmentation level provided the highest BLEU score and the lowest perplexity, suggesting that this method was particularly effective for this pair. For the fr-fon and fr-wol pairs, similar trends were observed. In both cases, out-lang switchout achieved the highest BLEU scores while sentence concatenation with backtranslation produced the lowest perplexity values. These complementary improvements highlight that the optimal augmentation strategy may differ even within related language pairs. A detailed result for each augmentation technique can be seen in Table \ref{tab:sentenceconcat_results} and Table \ref{tab:switchout_results}. \\
Overall, our results show that data augmentation techniques can greatly improve machine translation performance in settings with few resources. However, the benefits depend on the language pair, the level of augmentation used, and the type of augmentation chosen. In some instances, switchout outperformed sentence concatenation with backtranslation (as seen in en-hau, en-yor, fr-fon, and fr-wol), while in others (such as en-tsn), sentence concatenation yielded superior results. Notably, for en-swa, characterised by abundant parallel data, the baseline model remained the best, indicating that the impact of augmentation may be less pronounced when ample training data is available.

\begin{table}[htbp]
\centering
\caption{Average results for the data augmentation techniques used on a machine translation task across 6 languages. The best result for each language category is highlighted.}
\label{tab:data augmentation_results}
\scriptsize
\begin{tabular}{llrrlrrrrrrrrrrr}
\toprule
\textbf{Language} & \textbf{Aug. Technique} &  \textbf{Technique Type} & \textbf{Aug. Percentage} & \textbf{Bleu} &  \textbf{Loss} & \textbf{Parallel samples}\\
\midrule

        en-hau &  Baseline &  & 0 & 5.1028 & 2.8754 & 5865 \\
               &  Switch Out &In-lang & 50 &  \textbf{9.6464} & 2.5301 & 8797\\
               &  Sentence Concat & & 10 &  8.9139 & \textbf{2.5187} & 12316 \\
        \midrule 
        en-swa &  Baseline & & 0& \textbf{25.7951} & \textbf{1.8199} & 30782\\
               &  Switch Out & Out-lang & 10 & 25.7406 & 1.9711 & 33860 \\
               &  Sentence Concat &  & 10 & 25.1339 & 2.2478 & 64641\\
        \midrule    
        en-tsn &  Baseline & & 0 & 4.1371 & 2.8710 & 2100 \\
               &  Switch Out  & In-lang & 100 &  10.1873 & 2.6764 & 4200\\
               &  Sentence Concat &  & 20  & \textbf{11.8206} & \textbf{2.5347} & 4620\\
        \midrule       
        en-yor &  Baseline & & 0 & 6.4219 & 2.0800 & 6644 \\
               &  Switch Out & Out-lang & 30 & \textbf{9.1402} & \textbf{1.9883} & 8637 \\
               &  Sentence Concat & & 40 & 7.5838 & 2.1839 & 15845\\
        \midrule      
        fr-fon &  Baseline & & 0 & 0.8853 & 4.8712 & 2637\\
               &  Switch Out & Out-lang & 50 & \textbf{3.4523} & 2.9817 & 3955 \\
               &  Sentence Concat & & 10 & 2.7834 & \textbf{2.7651} & 5537 \\
        \midrule      
        fr-wol & Baseline     & & 0 &  2.0927 & 6.0191 & 3360\\
               & Switch Out & In-lang & 100 & \textbf{7.7010} & 3.2954 & 6720 \\
               & Sentence Concat & & 20 & 7.0803 & \textbf{2.8596} & 7392 \\
\bottomrule
\end{tabular}
\end{table}

\begin{table}[htbp]
\centering 
\caption{Comprehensive results for the sentence concatenation with back translation data augmentation used across 6 languages.}
\label{tab:sentenceconcat_results}
\scriptsize
\begin{tabular}{lllllll}
\toprule
\textbf{Language} &  \textbf{Aug. \%} & \textbf{Bleu} & \textbf{Loss} & \textbf{Parallel samples}\\
\midrule

        en-hau & 10 & 8.9139 & 2.5187 & 12316  \\
               & 20 & 7.2623 & 2.6221 & 12903 \\
               & 30 & 7.7952 & 2.6087 & 13489 \\
               & 40 & 7.6220 & 2.5645 & 14076 \\
        \midrule 
        en-swa & 10 & 25.1339 & 2.2478 & 64641  \\
               & 20 & 24.4418 & 2.2771 & 67719 \\
               & 30 & 24.6648 & 2.3135 & 70792  \\
               & 40 & 23.4520 & 2.4037 & 73873 \\
        \midrule    
        en-tsn & 10 & 8.6216 & 2.9135 & 4410 \\
               & 20 & 11.8206 & 2.5347 & 4620 \\
               & 30 & 9.6371 & 2.5540 & 4830  \\
               & 40 & 10.5753 & 2.5622 & 5040  \\
        \midrule       
        en-yor & 10 & 7.2755 & 2.1021 & 13952 \\
               & 20 & 7.4992 & 2.1161 & 14616 \\
               & 30 & 3.9273 & 6.7230 & 15281 \\
               & 40 & 7.5837 & 2.1840 & 15845 \\
        \midrule      
        fr-fon & 10 & 2.7834 & 2.7651 & 5537 \\
               & 20 & 2.7703 & 2.7205 & 5801 \\
               & 30 & 2.0730 & 3.6220 & 6065 \\
               & 40 & 2.7830 & 2.8125 & 6328 \\
        \midrule      
        fr-wol & 10 & 6.6867 & 2.7980 & 7056 \\
               & 20 & 7.0802 & 2.8596 & 7392 \\
               & 30 & 7.0072 & 2.8883 & 7728 \\
               & 40 & 7.0292 & 2.9181 & 8064 \\
\bottomrule
\end{tabular}
\end{table}
\clearpage
% \begin{table}[htbp]
\begin{table*}[t]
\centering
\caption{Comprehensive results for the two types of switch-out data augmentation used across 6 languages.}
\label{tab:switchout_results}
\scriptsize
\begin{tabular}{lllllllll}
\toprule
\textbf{Language} &  \textbf{Aug. \%} & \textbf{Bleu (In-lang)} &  \textbf{Bleu (Out-lang)} & \textbf{Loss (In-lang)} &  \textbf{Loss (Out-lang)} & \textbf{Parallel samples}\\
\midrule

        en-hau & 10 & 6.0781 & 2.8641 & 2.6794 & 7.5862 & 6451 \\
               & 20 & 2.7535 & 8.9679 & 2.9040 & 2.5346 & 7038 \\
               & 30 & 8.3668 & 8.6058 & 3.0087 & 2.5536 & 7624 \\
               & 50 & 9.6464 & 3.3819 & 2.5301 & 2.7050 & 8797 \\
               & 100 & 7.2026 & 7.4002 & 2.7043 & 4.0807 & 11730 \\
        \midrule 
        en-swa & 10 & 25.4653 & 25.7406 & 1.9726 & 1.9711 & 33860 \\
               & 20 & 25.5844 & 25.6335 & 2.0029 & 2.0044 & 36938 \\
               & 30 & 24.9896 & 25.5244 & 2.0332 & 2.0397 & 40016 \\
               & 50 & 24.9608 & 25.1808 & 2.1103 & 2.1166 & 46173 \\
               & 100 & 24.3488 & 24.4622 & 2.3137 & 2.3061 & 61564 \\
        \midrule    
        en-tsn & 10 & 7.7400 & 6.9780 & 2.6993 & 4.1369 & 2310 \\
               & 20 & 4.7364 & 9.7105 & 6.2975 & 2.7723 & 2520 \\
               & 30 & 9.842 & 3.4011 & 6.9919 & 3.1952 & 2730 \\
               & 50 & 4.4918 & 5.5807 & 2.7382 & 4.1337 & 3150 \\
               & 100 & 10.1873 & 9.5566 & 2.6774 & 2.5856 & 4200 \\
        \midrule       
        en-yor & 10 & 7.6303 & 8.962 & 2.0136 & 1.9711 & 7308  \\
               & 20 & 5.7969 & 7.7578 & 2.1545 & 2.0579 & 7972\\
               & 30 & 8.4158 & 9.1402 & 1.9870 & 1.9883 & 8637 \\
               & 50 & 8.4978 & 6.9424 & 2.0191 & 2.1557 & 9966 \\
               & 100 & 7.5750 & 5.5488 & 2.0784 & 2.8268 & 13288 \\
        \midrule      
        fr-fon & 10 & 3.2746 & 2.2964 & 2.8106 & 2.7403 & 2900 \\
               & 20 & 3.1829 & 2.9887 & 7.9939 & 2.8279 & 3164 \\
               & 30 & 3.0442 & 2.6116 & 2.8584 & 2.7712 & 3438 \\
               & 50 & 3.3083 & 3.4523 & 2.9417 & 2.9817 & 3955 \\
               & 100 & 2.2855 & 3.3356 & 3.1158 & 3.1754 & 5274 \\
        \midrule      
        fr-wol & 10 & 7.5748 & 6.2492 & 2.8215 & 2.8114 & 3696 \\
               & 20 & 5.1582 & 4.9001 & 3.0102 & 3.8348 & 4032 \\
               & 30 & 6.5973 & 6.8814 & 2.9199 & 2.8447 & 4368 \\
               & 50 & 7.1853 & 6.6625 & 2.9324 & 2.9850 & 5040 \\
               & 100 & 7.7010 & 6.245 & 3.2954 & 3.1230 & 6720  \\
\bottomrule
\end{tabular}
\end{table*}

\section{Conclusion \& Future Work}
In this study, we investigated the impact of two data augmentation techniques, switchout and sentence concatenation with BT, on machine translation tasks for low-resource African languages. Our findings indicate that these techniques can improve the performance of machine translation models across most language pairs, highlighting the potential of data augmentation in addressing the scarcity of labeled data and improving translation accuracy. Beyond improving translation accuracy, our study contributes to the broader goal of African language preservation and this is important given the historical marginalisation of African languages. \\

Future work will extend our investigation to larger models such as M2M100 and NLLB. We also plan on exploring generative augmentation techniques, including data generation using large language models, to further enhance translation performance in truly low-resource scenarios. These efforts will help establish robust translation frameworks tailored to the linguistic and data constraints of African languages, ultimately contributing to more inclusive and effective machine translation systems.

% \section*{Acknowledgments}
%  Remember to thank MLC

% \section{Citations and Bibliography}
% \label{sec:cite}

% \bibliography{arxiv}

\small
\bibliographystyle{plainnat}
\bibliography{arxiv}

% \clearpage  

% \appendix
% \section*{Appendix A}
% \addcontentsline{toc}{section}{Appendix A}

% Comprehensive results for the experiments carried out with two data augmentation techniques on 6 low-resource African languages.

\vspace{1em}

%\appendix \section{Comprehensive results for the experiments carried out with two data augmentation techniques on 6 low-resource African languages.}\label{apd:first}

\end{document}